\documentclass[letterpaper, 10 pt, conference]{ieeeconf}  % Comment this line out if you need a4paper

\IEEEoverridecommandlockouts                              % This command is only needed if 
\usepackage{graphics} % for pdf, bitmapped graphics files
\usepackage{subfigure}
\usepackage{epsfig} % for postscript graphics files
\usepackage{multicol}
\usepackage{amssymb}  % assumes amsmath package installed
\usepackage{algorithm}
\usepackage{algorithmicx}
\usepackage{algpseudocode}
\usepackage{soul}
\usepackage{comment}
\usepackage{epstopdf}
\usepackage{hyperref}
\usepackage{multirow}
\usepackage{cite}
\usepackage{amsmath}
\usepackage{hyperref}
\usepackage{array}
\usepackage{balance}
\usepackage[flushleft]{threeparttable}

\title{\LARGE \bf
	ABatRe-Sim: A Comprehensive Framework for \underline{A}utomated \underline{Bat}tery \underline{Re}cycling
\underline{Sim}ulation}

\author{
	Huanqing Wang,
        Kaixiang Zhang,
        Keyi Zhu,
        Ziyou Song,
	Zhaojian Li
	\thanks{Huanqing Wang, Kaixiang Zhang, Keyi Zhu and Zhaojian Li are with the Department of Mechanical Engineering, Michigan State University, East Lansing, MI 48824, USA (E-mail: \{wanghu26, zhangk64, zhukeyi1, lizhaoj1\}@egr.msu.edu).}% <-this % stops a space
 \thanks{Ziyou Song is with the Department of Mechanical Engineering, National University of Singapore, Singapore 117575, Singapore.
E-mail: ziyou@nus.edu.sg.}
}%

\begin{document}
	
	\maketitle
	\thispagestyle{empty}
	\pagestyle{empty}

	%%%%%%%%%%%%%%%%%%%%%%%%%%%%%%%%%%%%%%%%%%%%%%%%%%%%%%%%%%%%%%%%%%%%%%%%%%%%%%%%
	\begin{abstract}
		With the rapid surge in the number of on-road Electric Vehicles (EVs), the amount of spent lithium-ion (Li-ion) batteries is also expected to explosively grow. The spent battery packs contain valuable metal and materials that should be recovered, recycled, and reused. However, only less than 5\% of the Li-ion batteries are currently recycled, due to a multitude of challenges in technology, logistics and regulation. Existing battery recycling is performed manually, which can pose a series of risks to the human operator as a consequence of remaining high voltage and chemical hazards. Therefore, there is a critical need to develop an automated battery recycling system. In this paper, we present ABatRe-sim, an open-source robotic battery recycling simulator\footnote{The simulation platform is open-source and accessible at \url{https://github.com/hwan30/ABatRe-Sim}.}, to facilitate the research and development in efficient and effective battery recycling automation. Specifically, we develop a detailed CAD model of the battery pack (with screws, wires, and battery modules), which is imported into Gazebo to enable robot-object interaction in the robot operating system (ROS) environment. It also allows the simulation of battery packs of various aging conditions. Furthermore, perception, planning, and control algorithms are developed to establish the benchmark to demonstrate the interface and realize the basic functionalities for further user customization. Discussions on the utilization and future extensions  of the simulator are also presented.

 % As the depletion of fossil fuels continues, the transition to electric vehicles (EVs) and hybrid electric vehicles (HEVs) has become increasingly important in recent years. However, this shift to EVs has created another issue: the battery packs contain toxic substances and valuable metals that require proper recycling. Due to the complexity of battery packs, there are few open-source simulation environments available for researchers to study this topic, therefore, it is beneficial to establish a thorough simulation framework for battery recycling initiatives.

        %In this paper, we present a simulation framework for battery recycling applications in ROS (Robot Operating System). The framework integrates the entire process of robotic vision and control. It also provides simulation methods for stacked or assembled objects. The goal of this framework is to provide an open-source platform for researchers to further study and develop the technology needed for efficient and effective battery recycling. The framework is a useful tool for evaluating and improving the efficiency of battery recycling processes, as well as exploring new solutions for the responsible disposal of used batteries.

	\end{abstract}

	%%%%%%%%%%%%%%%%%%%%%%%%%%%%%%%%%%%%%%%%%%%%%%%%%%%%%%%%%%%%%%%%%%%%%%%%%%%%%%%%
	\section{Introduction}
	
	%The increasing regulations from the EPA on fuel economy have driven the automotive industry towards electric vehicles (EVs) and hybrid electric vehicles (HEVs)\cite{8891382}, highlighting the need for proper battery recycling.
 The past decade has witnessed an explosive growth in Electric Vehicles (EVs) due to their environmental friendliness, contribution to energy security, and cost reduction in operation and maintenance. As a result, there has  been 
 a concurrent surge in the number of spent lithium-ion (Li-ion) batteries;  it is estimated that the amount of spent Li-ion batteries will hit 2 million metric tons per year by 2030 \cite{battery-number}. The spent battery packs contain valuable metals and materials (e.g., cobalt, nickel, lithium, and manganese) that account for more than half of the battery's cost and can (and should) be recovered, processed, and reused. 
 However, only a very small portion of batteries are recycled in a limited number of mass battery recycling plants \cite{battery-number}. These factories mostly rely on manual labor, and due to the remaining high-voltage of battery packs and chemical hazards, specialized training is required for workers. Moreover, the labor involved in battery recycling is often monotonous and repetitive, making it a prime candidate for robotic automation to improve the operation efficiency and address the issue of labor shortage. 
    
    Developing robotic systems for automated battery recycling, however, is a daunting task due to the great complexity of battery packs that involve various materials, wires, metals, and sensors. Those components come with various aging conditions and are inter-connected with complicated structures. Dexterous and precise motions are required in order to perform challenging operations such as pulling, unscrewing, and lifting to successfully dismantle the batteries without damaging the chemical packs in the battery cell. Furthermore, proper operation procedure needs to be established to minimize the battery pack damages. Despite tremendous interest, the development of robotic battery recycling is still in its infancy. Existing robotic systems under exploration are only designed for very specific tasks, such as unscrewing \cite{WEGENER2015716}, and gripping and cutting \cite{cutting}. A comprehensive robotic solution to recycling is very much lacking.

    Since a battery pack is generally costly and burdensome to acquire, maintain, and manage, there is a critical need for a comprehensive simulation platform to facilitate the development of enabling robotic battery recycling technologies, including perception, planning, and controls. Towards that end, several simulation platforms have been attempted. For example, \cite{robotics10020082} focuses on the sorting problem in battery recycling where battery parts are positioned flat on the ground and sorted by a manipulator.  Note that it is oversimplified  without considering the most labor-intensive part of the process - disassembly. In \cite{10.3389/frobt.2021.688275}, the battery pack is treated as a single object and a cutting procedure is simulated.  However, the simulation offers limited fidelity as the module remains unchanged visually throughout the simulation, inadequate to provide a realistic and high-fidelity representation of the battery systems. %Attempts have been made to create a Computer-Vision assisted manipulator framework\cite{9144984}, where researchers train their model for detecting specific objects and then send object coordinates to manipulator for pickup. However, this framework is not suitable for real-time implementation due to the use of built-in Gazebo services for target location information, which are not feasible in the real world. 
    It is noted that \cite{HELLMUTH2021398} analyzes the battery packs and provides insights on the methodology of the general disassembly process, which is useful for designing corresponding robotic procedures. The simulation of a custom bit-changing tool to unscrew bolts is also studied in \cite{7090386,WEGENER2015716}, which, however, only include one specific procedure (i.e., unscrewing) of the battery disassembly process. %Although these concepts could potentially inspire other researchers, they may not provide opportunities for advancement in previous work, as other researchers lack access to either a real battery pack or a simulation model. 
    In summary, despite some advances, a comprehensive simulation framework that can support the development of robotic battery recycling still falls short. 
    
     In this paper, we present a comprehensive simulation platform for robotic battery recycling studies. The goal is to  provide a versatile simulation tool for researchers and practitioners in the field to develop and refine processes and algorithms (e.g, perception, planning, and control) to enable efficient and effective robotic procedures. %and introduce methods for overall object detection, control, and simulation. Our objective is to provide a comprehensive solution and a valuable tool for researchers to improve battery recycling processes and explore new planning and control algorithms. The methods and techniques presented in this paper have the potential to impact the battery recycling industry.
    Towards the goal, we develop a simulation platform that treats the disassembly process of a generic EV battery pack consisting of 4 modules with inter-connected bolts and cables. We import the battery pack specifications from a Computer-Aided Drawing (CAD) model into Gazebo, enabling object interactions by exploiting its physics engine. Procedures such as unscrewing, pulling, and lifting are included to simulate the essential tasks that arise in the disassembly process. Furthermore, a default robot arm and sensor set is provided with benchmark perception, planning and control algorithms to demonstrate the interface and offer basic functionalities for further extensions. Overall, the  open-source platform provides an efficient yet representative simulation of the robotic disassembly process, which can be used to develop, evaluate, and test  relevant algorithms. %Those three components are selected for demonstration for the following reasons. Unscrewing bolts is the most repetitive task in battery disassembly. High-voltage copper cables connect the modules, and both ends of the cables are secured with bolts. A valid transition during removal between these two is of great significance. Heavy battery modules are best lifted by machine. Other parts, such as the Battery Management System (BMS) module, which require pressing down extremely small and delicate tabs on the electrical socket, have not been investigated in the simulation for practical purposes since there are almost impossible to simulate.

    The rest of this paper is organized as follows. Section II introduces the development of the simulated hardware and environment while the benchmark perception, planning and control algorithms are presented in Section III to demonstrate the interface. The utilization and extendability of the simulation framework is discussed in Section IV. Finally, conclusions are drawn in Section V.  
    
    %The rest of the paper is organized as follows. Section II introduces the simulation platform development, presenting the creation of simulated hardware and environment. Section III discusses benchmark algorithm implementation, which explains how our perception and planning works, Section IV explains extensions and discussions, offering insight into how users can adapt our framework to their specific requirements. Finally, Section V comprises the conclusion and future work of this paper, summarizing our work and outlining our future plans for this project.

    	\begin{figure}[!h]
		\centering
		\includegraphics[width=0.9\linewidth]{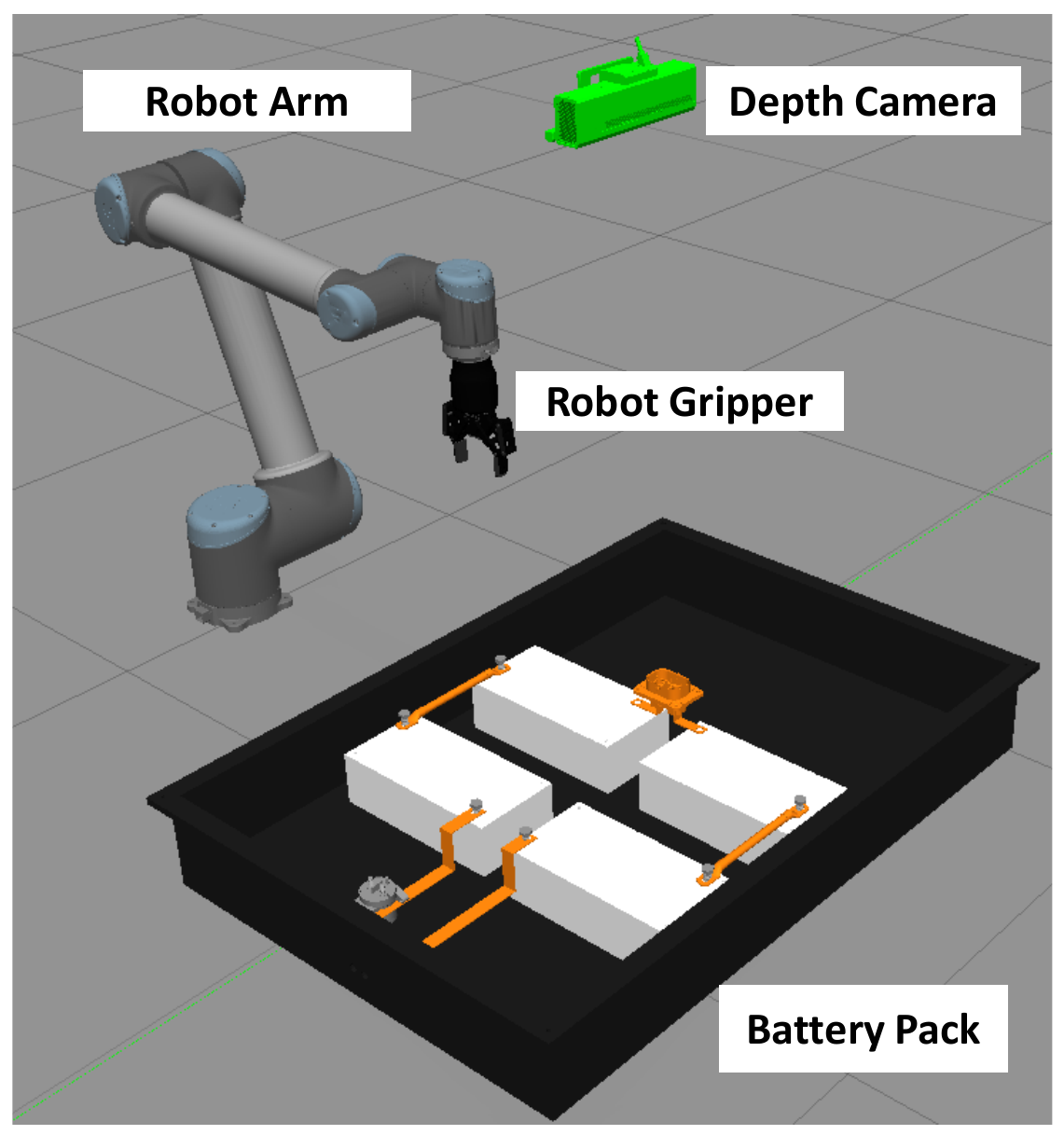}
		\caption{Illustration of the simulation environment with a battery pack, a robot arm/end-effector for manipulation, and a depth camera for perception.}\label{fig_Environment}
	\end{figure}

	\section{Simulation Platform Development} \label{sec_development}
	The major components of the developed simulation platform are depicted in Fig.~\ref{fig_Environment}, which includes a battery pack model, a sensor set for perception, and robot arm and end effector for manipulations. More specifically, the battery pack model is created as a CAD drawing and imported into Gazebo as an interactive object. The perception sensor provides  measurements (e.g., RGB image and 3D point cloud) necessary for perceiving the scene to guide robot operations. Lastly, a manipulator -- a combination of robot arm and end effector -- is used to realize robotic operations to disassemble the parts in the battery pack. We next present the details of each module in the following subsections.% The simulated hardware consists of a robot arm, robot gripper, and depth camera. The RGB-D camera captures real-time images, which are then sent to the object detection function for processing, while the robot arm and gripper work together to generate manipulations to disassemble the battery pack. The detailed creation process of the interactive object, as well as the relative functions of the hardware, will be presented in the following subsections.

	\subsection{Battery pack modeling} \label{sub_battery}
Despite the growing interest in robotic battery recycling, a suitable model integrated in robotic operating system (ROS) is still lacking. The CAD drawings of battery packs are considered proprietary information by the EV companies and are generally not available to the research community. The complexity and technical details make it challenging to create such drawings. Furthermore,  importing such complex models into ROS-based simulation environment presents a challenge, as the complexity of these models can lead to unexpected behaviors in the physics engine of simulators such as Gazebo \cite{Buehler2016}. While some sorting tasks have been simulated in the Gazebo environment, they typically involve objects that are either lying flat on the ground \cite{robotics10020082} or imported as a single piece \cite{10.3389/frobt.2021.688275}. To our knowledge, a simulation platform involving stacked or assembled battery objects is largely unavailable.

\begin{figure}[!h]
		\centering
		\includegraphics[width=0.98\linewidth]{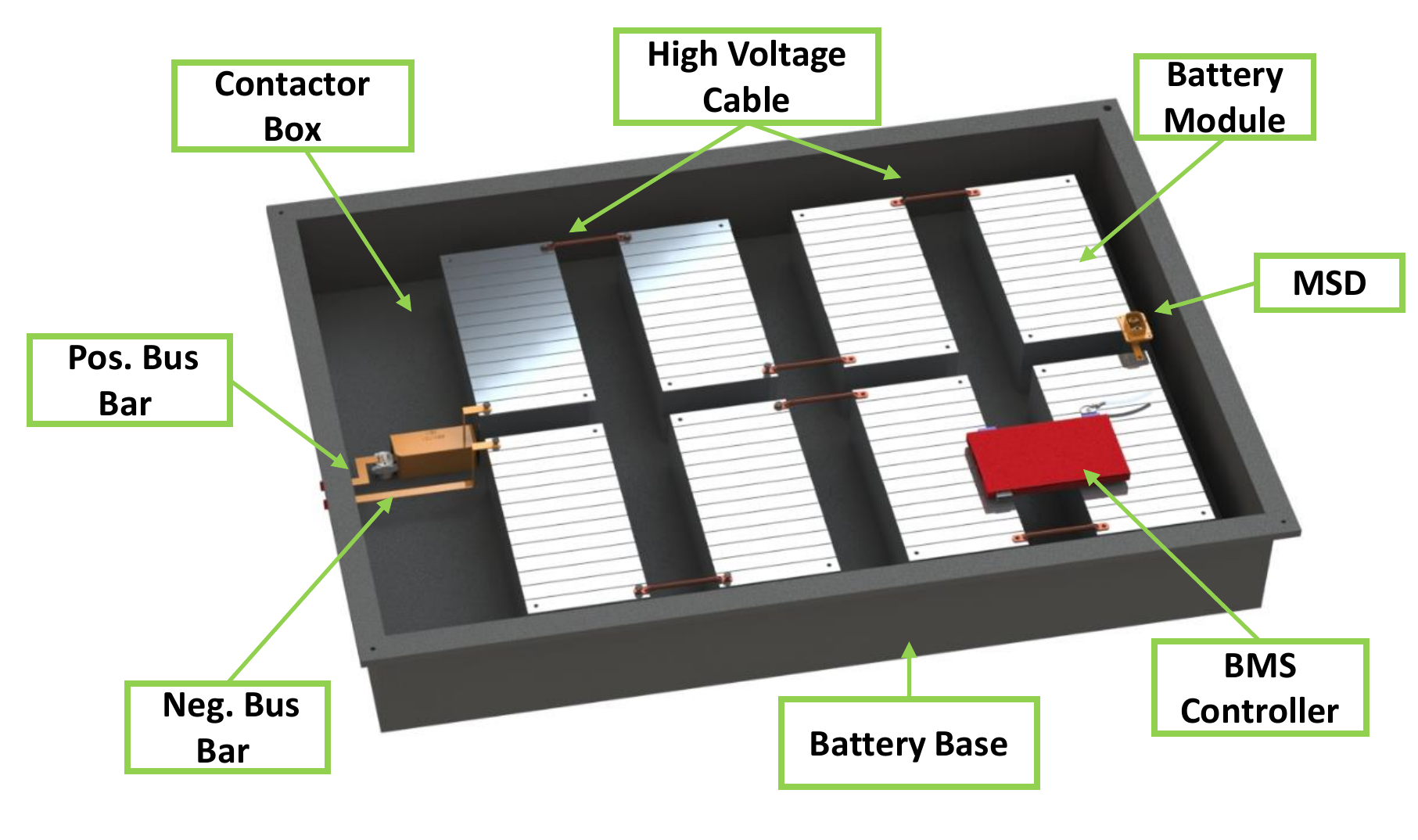}
		\caption{The battery management system modeled in SolidWorks.}\label{fig_bms}
	\end{figure}

To address the aforementioned challenges, a detailed battery model is created in SolidWorks with major components necessary for robotic disassembly.  As shown in Fig.~\ref{fig_bms}, the silver blocks represent battery modules, connected by high voltage cables in orange. The switch on the right side (orange) is Manual Service Disconnect (MSD), which, when unplugged, breaks down the pack voltage by half. The orange box on the left side is the contactor box, which contains high-voltage relays (contactors). The two orange lines on the left-hand sides are the positive bus bar and the negative bus bar, respectively. The red box is the BMS controller which is the ``brain'' of the battery pack (marked red only for demonstration purpose). The components mentioned above account for the major structural parts of the battery pack. %However, the Module control unit and wiring harness have not been modeled in this simulation. The wiring harness connects the Module control unit with the BMS module and transmits cell information to the main micro-controller. 
Since adding more components of the same category only involves repetitive work, we designed a generic EV battery pack with four modules in Gazebo simulation, as shown in Fig.~\ref{fig_Environment}. This battery pack also includes the battery base, two cables, positive and negative bus bars, one contactor, one MSD, and multiple bolts.
    
    In this framework, we will demonstrate the detection and disassembly of three main classes: bolts, cables, and modules. The shape is imported through model editors. Its physics properties are further hand-modified in XML scripts to reduce the processing workload. Each component is set to static in its XML script, but will only be replaced with a movable version prior to picking due to the fact that fewer interacting parts will reduce the physics engine processing load. For example, movable bolts will be spawned prior to the unbolting action while static bolts will be simultaneously deleted. The same approach is applied to cables and modules. In this way, a seamless transition is achieved without noticeable visual differences. We use the built-in physics engine in Gazebo to simulate the object interactions. Specifically, we exploit the grasp plugin \verb|libgazebo_grasp_fix.so|~\cite{Buehler2016} to establish a connection between two components when the gripper approaches a particular position, and the object can be dropped from the gripper when it is open.

\begin{figure}[!h]
		\centering
		%\hspace{-55pt}
  \includegraphics[width=0.95\linewidth]{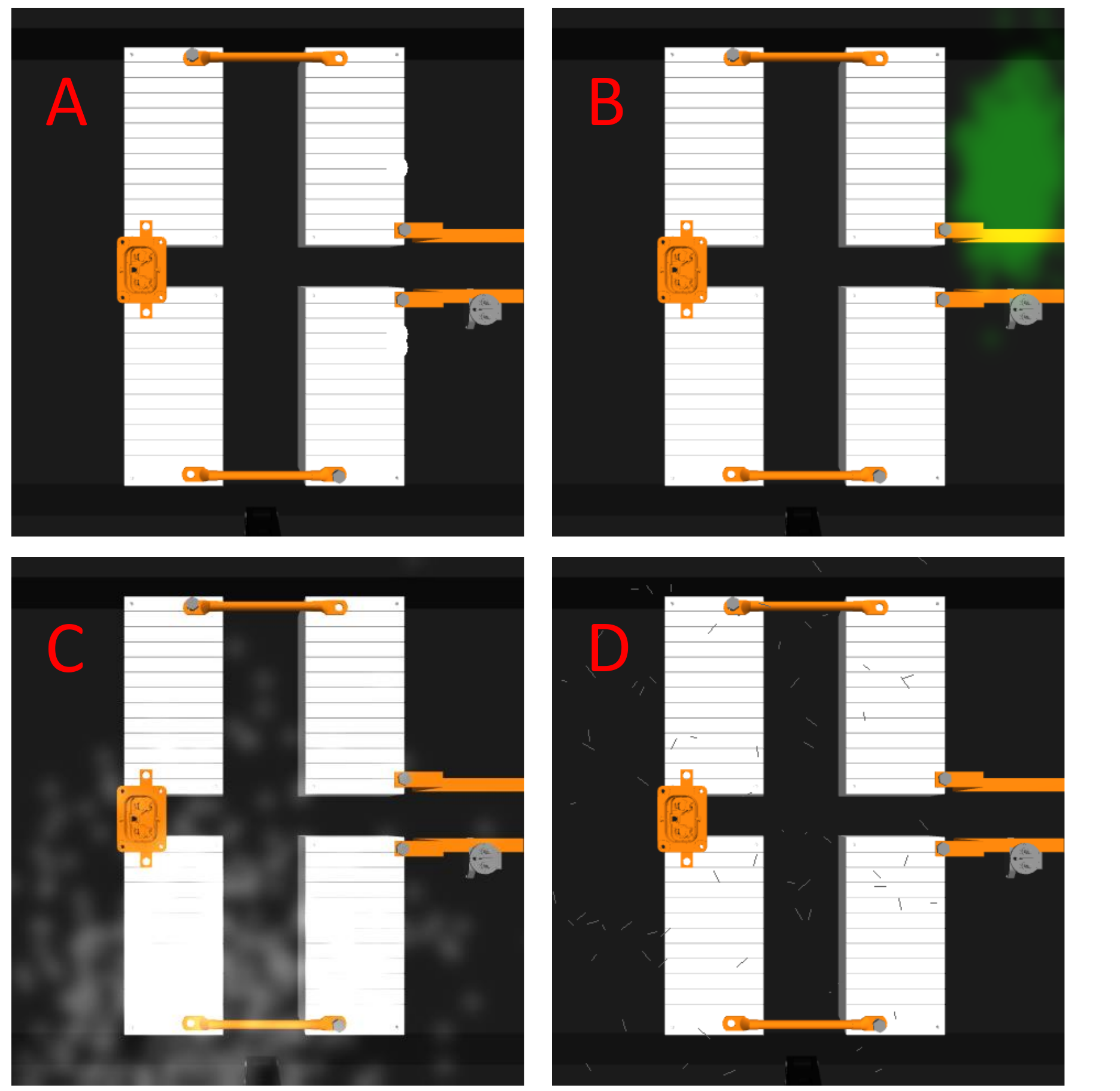}
		\caption{Function to generate simulated battery pack with  deformation (A), contamination (B), dust (C) and scratches (D). }\label{fig_Conditions}
	\end{figure}

   We have also taken into account scenarios where the battery pack is received damaged or in an imperfect condition, which could be due to internal failure, transportation damage, or aging. To address these imperfections, we have created image filters that are available to simulate such conditions.  Fig.~\ref{fig_Conditions} illustrates four simulated conditions that could occur when a pack is received at the disassembly station. Image A depicts a dent or deformation that could occur after the pack is pressed by other objects, resulting in convex shapes on the right edges of the two packs on the right. Image B simulates the case of an old improperly maintained pack where coolant leaked out, leaving behind corrosion and green coolant residuals. Image C simulates a long-sitting pack where a light layer of dust is present on top. Image D simulates the result of careless disassembly, where the pack is left with multiple marks on the metal and paint surfaces. These scenarios account for the possibility of a poorly maintained or transported battery pack. We plan to continue adding more scenarios to improve the capability of simulating real-world situations.

	\subsection{Robot manipulator and sensing modality} \label{sub_robot}
\begin{figure}[!h]
		\centering
		\includegraphics[width=0.9\linewidth]{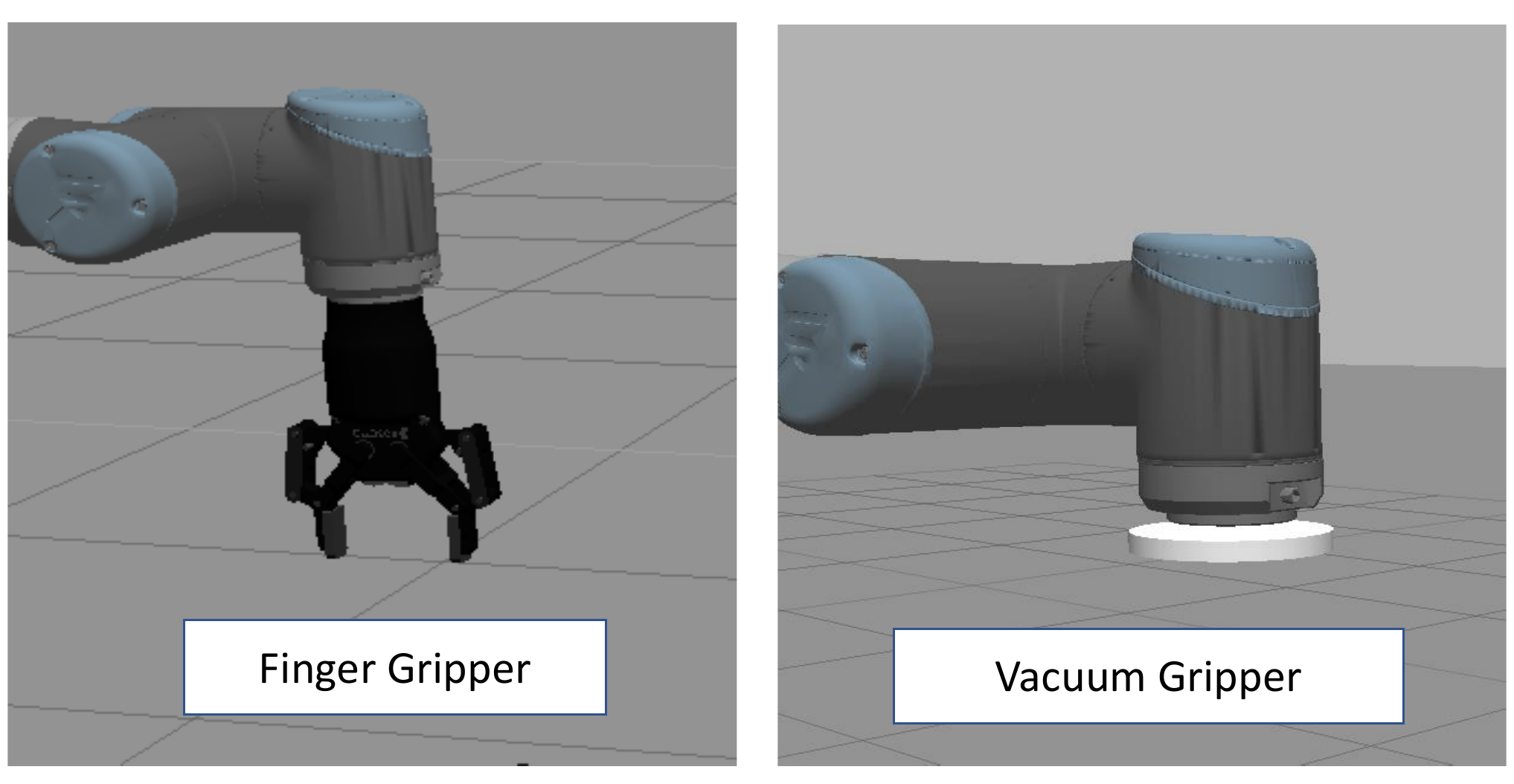}
		\caption{Two types of grippers are used in our simulation.}\label{fig_Grippers}
	\end{figure}

         While a number of robot arms can be used for this application, we use Universal Robot 10,  a 6-axis robotic arm, as the default in the simulation platform. Note that the user can choose other robotic arms in the simulation (see Section IV for more details). In ROS, robots are described using Unified Robot Description Format (URDF) files, which describe the robot's joint structures and kinematic relations. The files can include additional details such as the visual and collision properties of links, joints, sensors, etc. XACRO files (XML Macros) are extensions of URDF that allow for smarter and more modular descriptions of robots. Mesh files provide 3D models for the visual representation of robots in simulations. These files are often in STL or Collada format. The package for the robot is referenced from the Universal Robot Github Page~\cite{rosindustrialur}. Moreover, as shown in Fig.~\ref{fig_Grippers}, two types of grippers are incorporated with the robot arm to facilitate the battery disassembly. Specifically, a Robotiq gripper~\cite{rosindustrialrobotiq} is utilized for the first two operations, which involve unscrewing the bolts and removing them as well as lifting the cables. Following the removal of these parts, a vacuum gripper is employed to pick up the modules. The vacuum gripper is driven by the \verb|libgazebo_ros_vacuum_gripper.so|, which essentially creates linkages between objects when a service is called. The animation  resembles that of a vacuum gripper. The robot arm and gripper are connected by creating an additional link in-between. The robot and gripper files are open-source ROS packages on GitHub~\cite{rosindustrialur,rosindustrialrobotiq}. Inspirations from other researchers' integration with Universal Robots ~\cite{Yang2018, Huang2018} also contribute to this project.

Furthermore, a sensor is needed to understand the scene, i.e., identifying and localizing the position of the objects. To that end, we adopt Kinetic RGB-D camera as the sensor modality, which is placed on top of the battery pack, as depicted by the green object in Fig.\ref{fig_Environment}. In Gazebo, the Kinect has a ready plugin that renders color and depth images through ROS topics. In our node, the color image is streamed into the object detection node for inferencing different battery parts (see Section III-A) and the depth image is then used to generate the target positions to guide the manipulations. 

\section{Benchmark Algorithm Implementation} \label{sec_benchmark}
\begin{figure}[!h]
		\centering
		\includegraphics[width=0.9\linewidth]{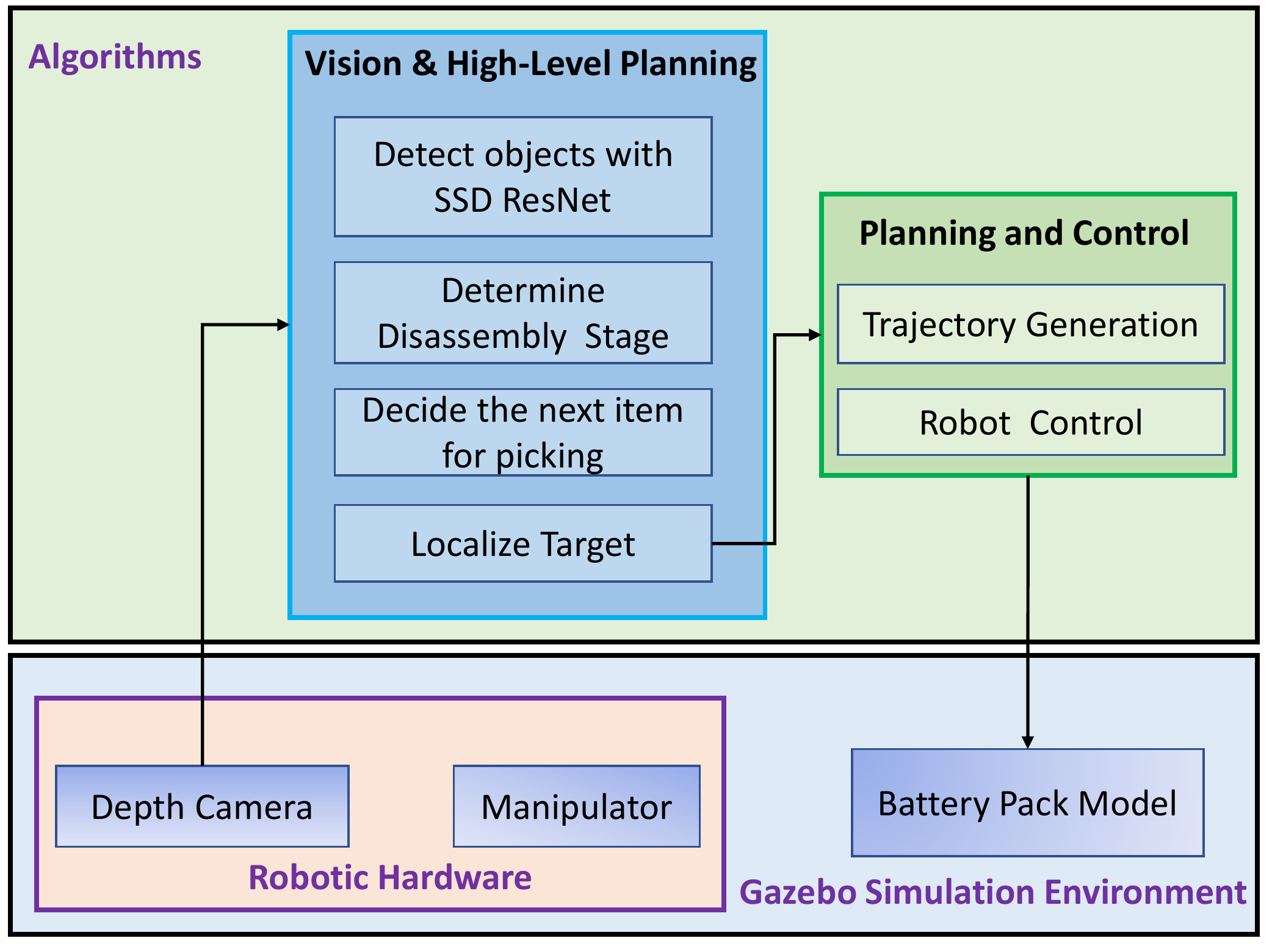}
		\caption{The interface between algorithm software and robot hardware.}\label{fig_Flow}
	\end{figure}

In this section, we describe the benchmark perception, planning, and control algorithms that come with the simulation platform, which demonstrates how to interface the simulation modules described in Section II and provides basic functionalities for customized extensions. 
Fig.~\ref{fig_Flow} depicts the interaction between the software and the simulated hardware. %The simulated hardware components include the robot and depth camera, which are  discussed in the earlier chapter and coexist in the gazebo simulation environment with our created battery pack model.
Specifically, RGB images from the camera are consumed by the perception node, and a deep learning algorithm is employed to detect various parts of the battery pack (e.g., screws, cables, battery module). A high-level task planner is then used to determine the disassembly stage and subsequently the next item to remove. The target $XYZ$ position is then retrieved from the depth image, which is subsequently sent to the low-level planning and control algorithms for low-level disassembly executions. We next present the details of the algorithms in the following subsections.%The image topics sent are processed by the perception node, which receives streamed images from the Gazebo camera and performs object detection inferencing. All of this process information, including stage flags and coordinates, is then integrated into a single Python script, which is the Vision Algorithm. The coordinates of the target is then passed through frame transformation for planning and control, resulting in the generation of low-level disassembling maneuvers.  

        \subsection{Deep learning for object detection} \label{sub_object}
Detecting various parts of the battery pack is a key step to perform disassembly. Deep learning-based techniques have been widely used for object detection and have received significant attention in recent years.
%Object detection is a crucial area of research in deep learning and has garnered significant attention in recent years. 
There are numerous machine learning frameworks such as PyTorch and TensorFlow, and there are also open-source implementations that use pre-trained models for common objects with webcams. However, it should be noted that although there exist many Gazebo simulation integrations for object detection, fewer of them use more recent frameworks like TensorFlow 2 and recent learning algorithms like YOLOv5.

\begin{figure}[!h]
		\centering
		\includegraphics[width=1\linewidth]{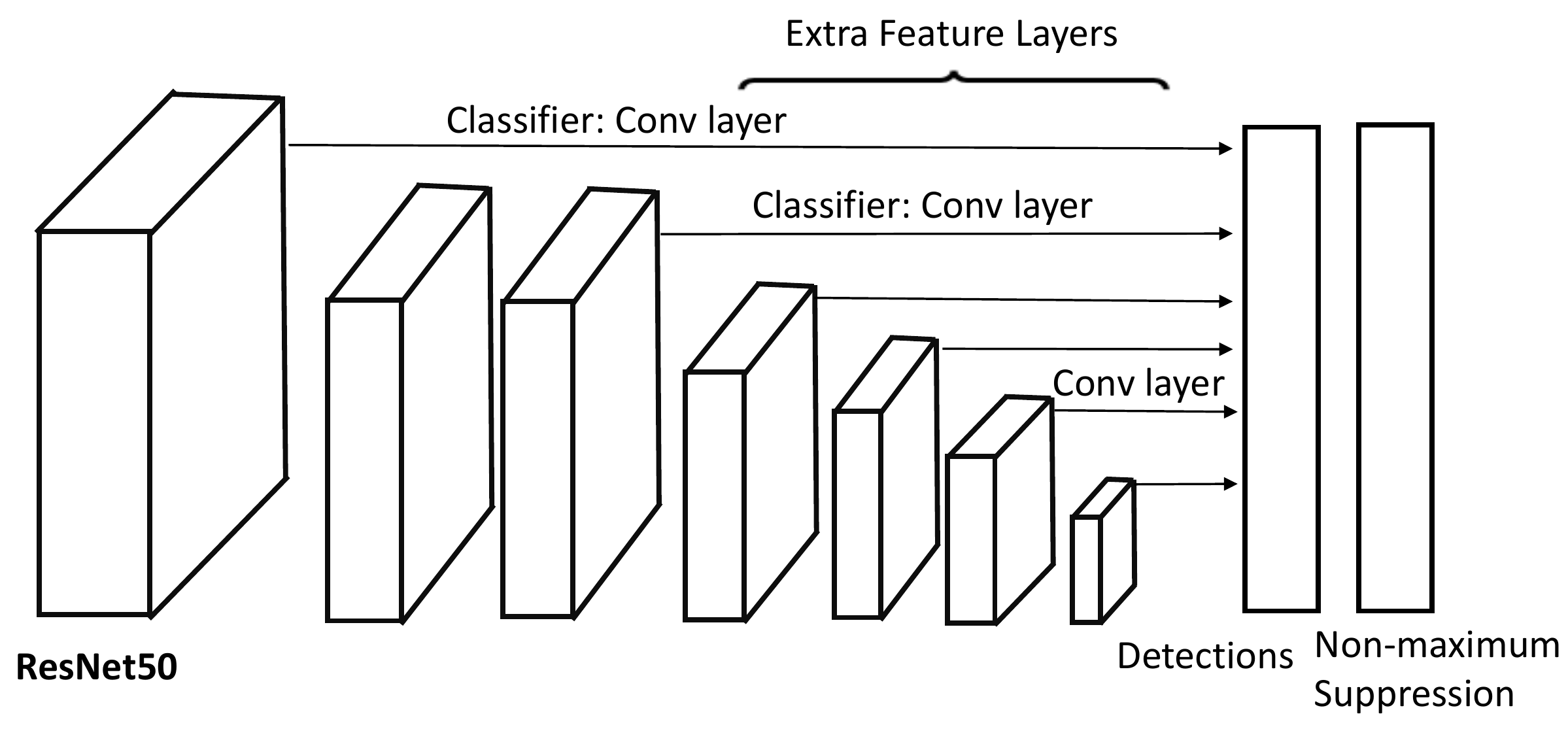}
		\caption{The architecture of SSD ResNet50 FPN.}\label{fig_CNN}
	\end{figure}

For the object detection task, we utilize TensorFlow 2.11.0 to deploy the learning algorithm. The model architecture is constructed with SSD ResNet50 V1 FPN 640x640\cite{tf}, which is selected from the open-source TensorFlow model zoo. 
%The model architecture is from the open-source TensorFlow model zoo - SSD ResNet50 V1 FPN 640x640\cite{tf}, which stands for Single Shot MultiBox Detector with ResNet-50 as its backbone and Feature Pyramid Network (FPN) for feature extraction. 
The model structure is shown in Fig.~\ref{fig_CNN}. Different from SSD~\cite{Liu_2016} that uses VGG16 as its backbone, the SSD ResNet FPN uses ResNet50 as the backbone and provides additional convolutional layers for classification~\cite{box}. ResNet50 is a deeper backbone than VGG16 as it has 50 layers and it is introduced to address the problem of vanishing gradients in deep neural network~\cite{https://doi.org/10.48550/arxiv.1512.03385}. The architecture also utilizes Feature Pyramid Network (FPN) for better feature extraction. FPN is a type of multi-scale feature extraction method that enhances the detection of objects of varying scales and sizes~\cite{https://doi.org/10.48550/arxiv.1612.03144}.

\begin{figure}[!h]
		\centering
            %\hspace{-45pt}
		\includegraphics[width=1\linewidth]{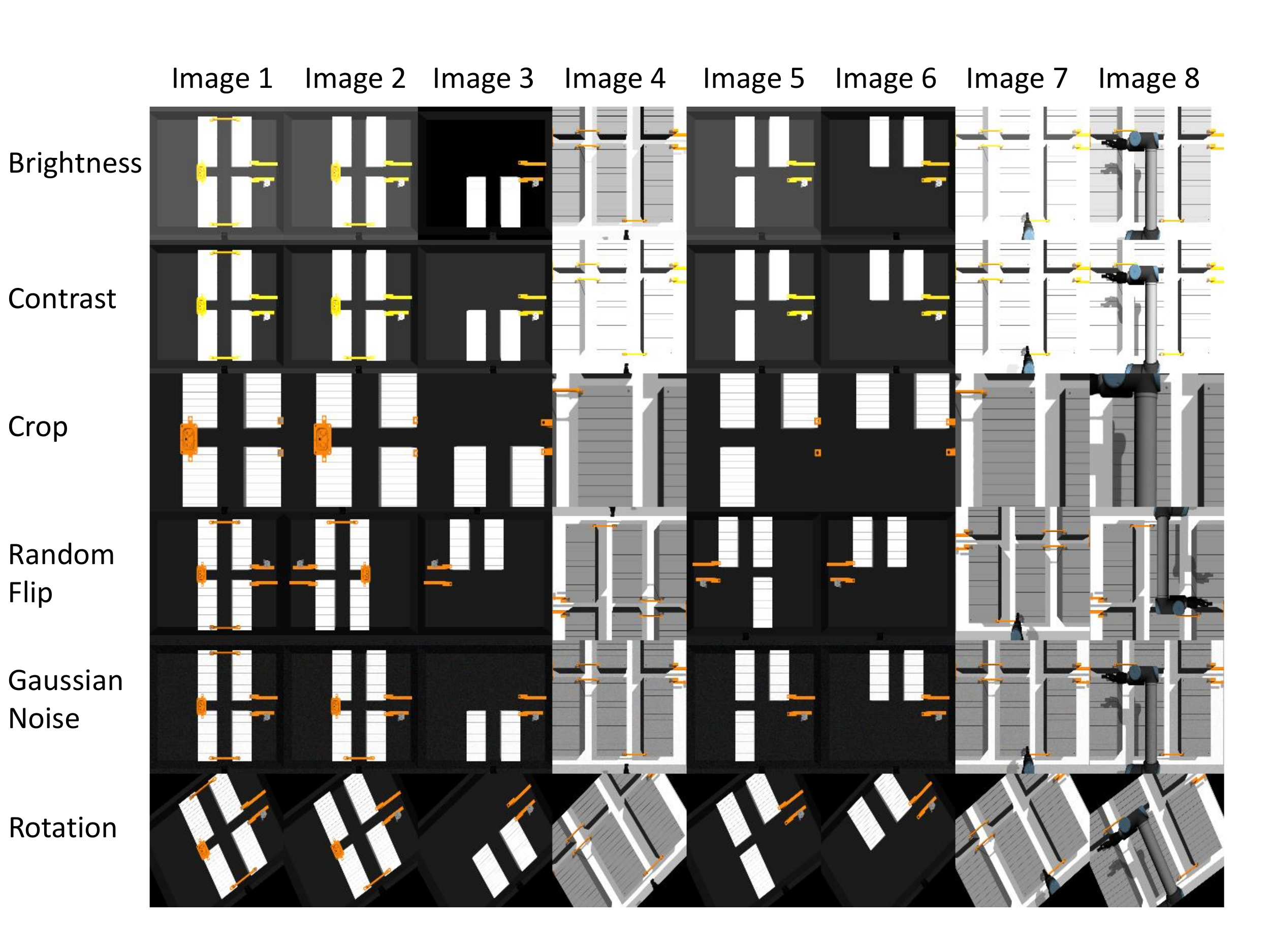}
		\caption{The dataset can be expanded to 6 different variations using data augmentation method.}\label{fig_data_aug}
	\end{figure}

 The task of image labeling is known to be a laborious process, requiring significant investments in time and human resources. Our research has determined that a dataset consisting of 120 labeled images is sufficient for our model to detect all three classes in a simulated environment, which exhibits lower complexity than real-world images. For the effective deployment of this approach in real-world applications, a larger dataset, and a corresponding increase in the volume of labeled images, will likely be needed. To account for different conditions in the real world, data augmentation techniques \cite{chen2022deep} is applied to extend the dataset, enabling tests under a broader range of conditions. Specifically, a data augmentation script is included in the simulation framework to characterize more variations in the data and improve the robustness of the models trained on the dataset. The tool can enlarge the dataset with random brightness, contrast, crop portion, flip orientation, Gaussian noise level, and rotation position. A sample of the enlarged dataset, consisting of eight distinct pictures, is shown in Fig. \ref{fig_data_aug}.
 
 The model training is performed on a separate  computer with  Ubuntu 20.04LTS operating system, an Intel Core i9-9820x CPU, and two NVIDIA GeForce RTX 2080Ti GPUs. In our training process, we specified the number of classes to be 3, the batch size to be 4, the warm-up steps to be 2000, and the training steps to be 25000. The entire training process took 1 hour 15 minutes and 9 seconds. Once the training is complete, the model is exported as a benchmark model for object detection purposes. The following are the final training metrics: classification loss of 0.03157, localization loss of 0.01083, regulation loss of 0.01415, and total loss of 0.05655. The exported model can then be employed for inferencing objects of interest. The inference node and Gazebo simulation were experimented on a Dell Precision PC with an Intel i7-11700 CPU and no dedicated GPU, and we only experience a small amount of latency.

The streaming of Kinetic camera images, object detection inference, and the communication of the process information (stage flags, coordinates) are all integrated into a single Python script called Vision Algorithm. The pseudo code for this module is shown in Algorithm 1, which processes the image and generates the bounding boxes with corresponding pixel coordinates. The pixel coordinates are transformed into the coordinates under the world frame, which are then passed to the planning and control module.
%The pixel coordinates are then subjected to a series of frame transformations before being sent to the planning and control algorithm. 

 \begin{figure}[!h]
		\centering
		\includegraphics[width=0.9\linewidth]{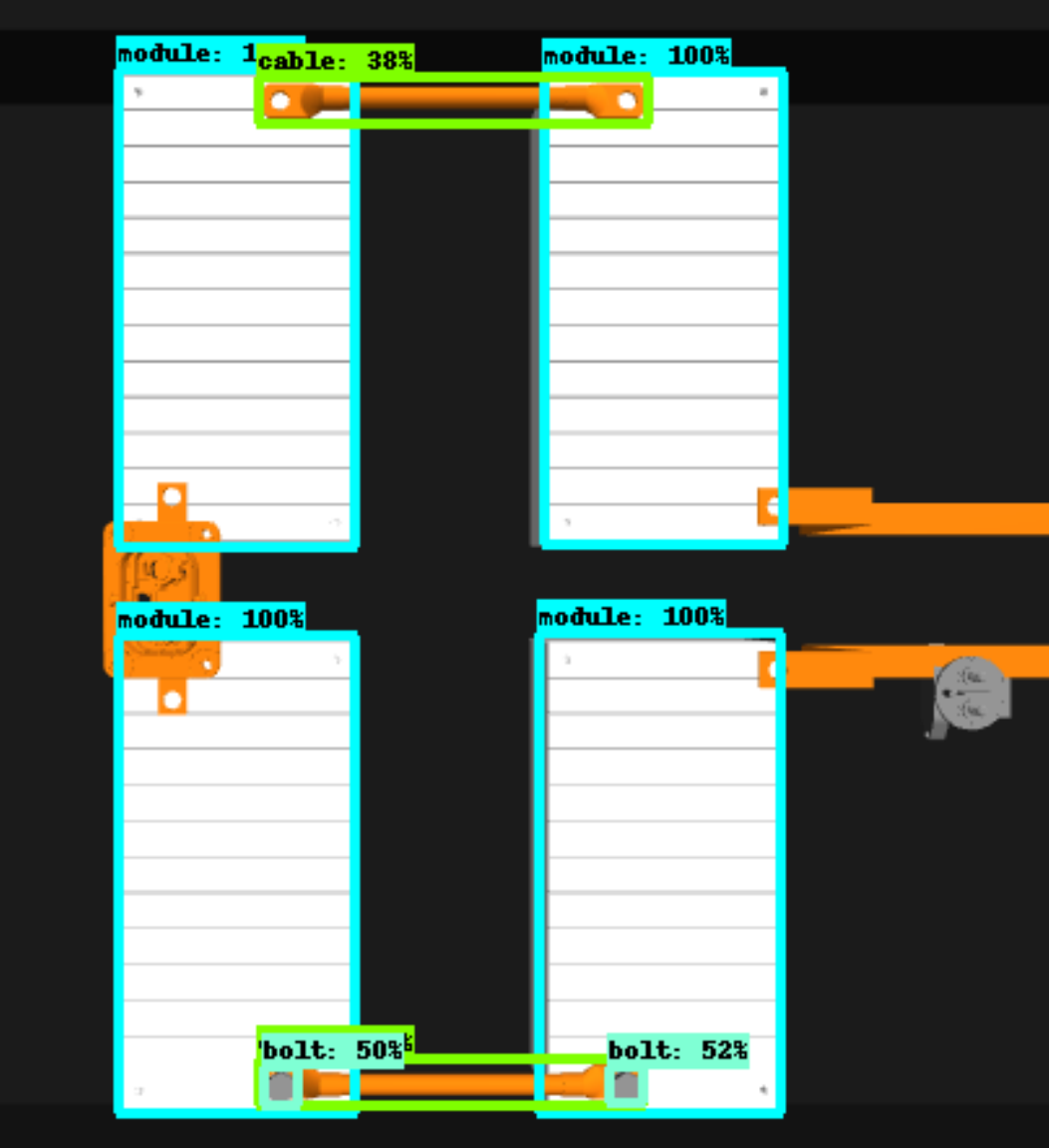}
		\caption{Results of objects of interest detected using the TensorFlow object detection API, with bounding boxes and confidence levels displayed.}\label{fig_object_dection}
	\end{figure}

\begin{algorithm}
\caption{Perception Algorithm}
\label{algorithm1}
\begin{algorithmic}[1]
\Procedure{Object Detection with Gazebo Depth Camera}{}
\State \textbf{Input:} Kinect camera images 
\State \textbf{Output:} Stage flag; Detected objects with bounding boxes; Image coordinates of Detected objects
\State Convert ROS image topics to OPENCV format.
\State TensorFlow inference model detects objects and creates the bounding boxes.
\State Initialize and append elements to bolt list if "Detection Class"=bolt. 
\State Initialize and append elements to cable list if "Detection Class"=cable.     
\State Initialize and append elements to module list if "Detection Class"=module.    

    \State Default: Stage flag = 1 ; Flag for bolt removal operation; Find the object center.
    \If{length(bolt list)==0}
        \State Stage flag = 2 (Flag for cable removal operation)
        \State Find the object center.
    \EndIf
    \If{length(cable list)==0}
        \State Stage flag = 3 (Flag for module removal operation)
        \State Find the object center.
    \EndIf

\State Return output
\EndProcedure
\end{algorithmic}
\end{algorithm}

\begin{algorithm}
\caption{Planning and Control Algorithm}
\label{algorithm2}
\begin{algorithmic}[2]
\Procedure{MoveIt Planner Battery Disassembly Process}{}
\State \textbf{Input:} Stage flag; World coordinates of detected objects $X, Y, Z$
\State \textbf{Output:} None
\State Initialize environment and planning scene.
\While{Stage flag == 1}
    \State Arm group planning - Move to target $(X, Y, Z)$
    \State  Gripper group planning - Close
    \State  Arm group planning - Twist motion
    \State  Arm group planning - Move to drop location
    \State  Gripper group planning - Open
    \EndWhile
    
\While{Stage flag ==2}
    \State Arm Group Planning - Move to target $(X, Y, Z)$
    \State  Gripper Group Planning - Close (ROS Service: Grasp Plugin $\rightarrow$ Attach)
    \State  Arm Group Planning - Move to drop location
    \State  Gripper Group Planning - Open (ROS Service: Grasp Plugin $\rightarrow$ Detach)
    \EndWhile
    
\While{Stage flag ==3}
    \State Arm Group Planning - Move to target $(X, Y, Z)$
    \State  ROS Service: Vacuum Gripper Plugin $\rightarrow$ On
    \State  Arm Group Planning - Move to drop location
    \State  ROS Service: Vacuum Gripper Plugin$ \rightarrow$ Off
    \EndWhile
    
\State Return output
\EndProcedure
\end{algorithmic}
\end{algorithm}

\begin{algorithm}
\caption{Editable Planner}
\label{algorithm3}
\begin{algorithmic}[3]
\Procedure{Biased RRT}{}
\State \textbf{Input:} Target pose in the Cartesian space $p_t$, Current joint configuration $q_0$
\State \textbf{Output:} A feasible joint trajectory $q=q_0q_1q_2...q_n$ 
\State Specify max iteration time $i_{max}$;
\State $q_t=InverseKinematics(p_t)$;
\If{$~IsValid(q_t)$}
    \State Return $q=\emptyset$;
\EndIf
\State $v_0=[q_0,\,d_0]$, where $d_0=0$;
\State $V=\{v_0\},\,E=\emptyset$;
\State $i=0$;
\While{$i<i_{max}$}
    \State $q_s=Sample(q_t)$;
    \State $q_{nearest}=FindNearest(q_s,V)$;
    \State $q_{new}=Steer(q_s,q_{nearest})$;
    \If{$~IsValid(q_{new},q_{nearest})$}
        \State Continue;
    \EndIf
    \State $d_{new}=d_{nearest}+||q_{new}-q_{nearest}||$;
    \State $v_{new}=[q_{new},d_{new}]$;
    \State $V=V\bigcup v_{new},\,E=E\bigcup(v_{nearest},v_{new})$;
    \State $v_{neighbors}=FindNeighbors(v_{new},V)$;
    \State $E=Rewire(q_{new},q_{neighbors},V,E)$;
    \If{$q_t\in V$}
        \State Return $q=GenTraj(V,E,q_t)$;
    \EndIf
    \State $i=i+1$;
\EndWhile
\State Return $q=\emptyset$
\EndProcedure
\end{algorithmic}
\end{algorithm}

	\subsection{Planning and Control} \label{sub_control}
	The planning and control algorithm of robotic motion is designed using the components from the open-source ROS package, \emph{MoveIt!}, and can be further modified by users with the functions provided in the demo. \emph{MoveIt!} is a comprehensive software framework for robotic motion planning and manipulation. Once a robot is set up and initialized in the \emph{MoveIt!} configuration, users can conveniently finish the task of planning and control for the robot through the API provided. By employing the \emph{Planning Scene Interface}, users can add the collision objects in the scene to achieve  collision-free planning. The arm (Arm group) and the gripper (Gripper group) are powered by MoveGroup, which is a component of \emph{MoveIt!} that provides interfaces for controlling motion planning and execution.
    In our demo, we exploit the Rapidly-exploring Random Tree (RRT)~\cite{lavalle2001randomized} algorithm for the planning of the robot as a C++ node in ROS, which makes it convenient for users to edit their own planning and control algorithms if necessary. With the Kinematics and Dynamics Library, we obtain a group of joint angles given a desired pose in the Cartesian workspace. Then we iteratively take samples in the joint configuration space and use the functions that \emph{MoveIt!} provides to check the validity of the sampled states. As the iteration progresses, a collision-free joint trajectory is then generated. When every point in the joint configuration space has the same probability of being sampled, this algorithm can be guaranteed successful if there is a feasible path. Taking some biased sampling strategy can accelerate the progress. Having obtained a feasible joint trajectory, we then use the Iterative Spline Parameterization algorithm from \emph{MoveIt!} to interpolate a timed trajectory that is executable for the robot. %At last, the robot would execute this trajectory. 

    The planning algorithm discussed above is outlined in Algorithm~\ref{algorithm3}. To briefly explain, Lines 4-11 initializes the algorithm. Line 5 obtains one group of joint configuration $q_t$ which complies with the given target pose $p_t$, and  the validity of $q_t$ is checked. Then a tree structure $\{V,\,E\}$ is initialized, where $V$ is the set of nodes, and $E$ is the set of edges. One node $v$ has two components, one being the joint configuration $q$, and the other being the distance to the root of the tree, i.e., $v_0$. In the while-loop, we first take the biased sampling, which has a probability of 0.2 for choosing $q_t$ as a result and 0.8 for uniform sampling. Then with the sampled $q_s$, the nearest node $q_{nearest}\in V$ is found. In Line 15,  $Steer$ function is used to replace $q_s$ with $q_{new}$ if $q_s$ is too far away from or too close to $q_{nearest}$, in order to efficiently explore the joint configuration space and find a feasible trajectory. After verifying the validity of the edge $(v_{nearest},v_{new})$, $v_{new}$ is added to the tree $\{V,\,E\}$. In Line 22-23,  the neighboring nodes of $v_{new}$ is checked to see whether those nodes are closer to $v_0$, if their "father" node is changed to $v_{new}$. If $q_t$ is added to the tree, a valid trajectory is then obtained by back-propagation in the tree $\{V,\, E\}$ from $v_t$.

     	\begin{figure*}[!h]
		\centering
		\includegraphics[width=0.9\linewidth]{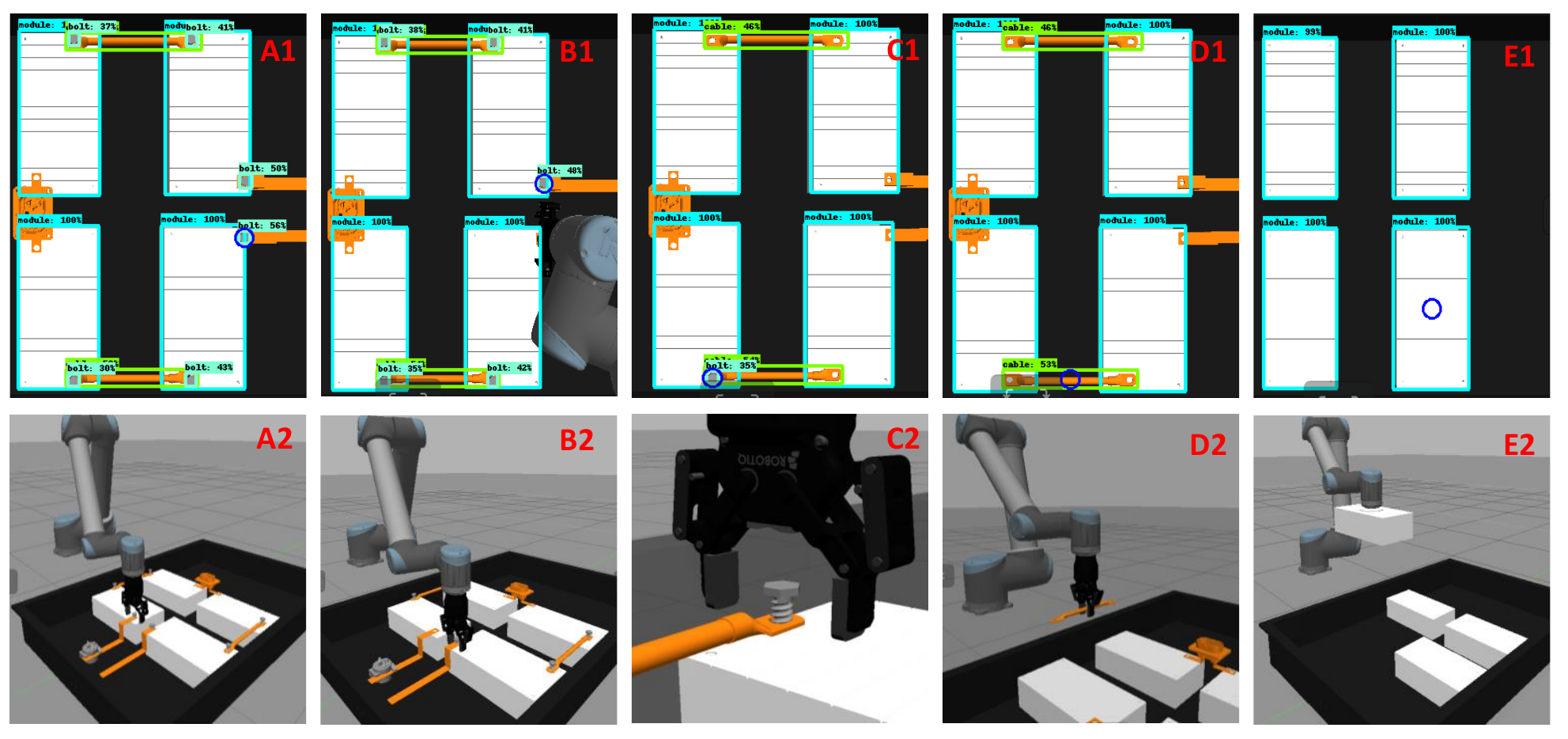}
		\caption{Illustration of the simulated battery disassembly process. A: The first bolt is detected by the perception node and the robot moves to execute its removal; B: The second bolt is then located by the perception node and the robot moves over to remove it; C: After several removals, the perception node detects the last bolt and the robot executes its removal; D: Once all bolts are removed, the perception node detects the cable and the robot proceeds to remove it; E: The detection of the module is carried out by the perception node, and the robot performs the removal action with a vacuum gripper.}\label{fig_Process}
	\end{figure*}

    In our simulator, high-level control actions, such as unscrewing bolts, removing cables, and lifting modules, are executed based on different stage flags as shown in Algorithm~\ref{algorithm2}, while the C++ node handles low-level trajectory generation and inverse kinematics. In order to initiate the simulation, it is crucial to launch all necessary functions, such as Gazebo world, frame transformation, vision, and control in ROS nodes, and establish communication through ROS topics. For the disassembling task, the first step is to unscrew the bolts. The detected bolts are appended to a list for preparation in the perception node. The first item in the list has its coordinates sent to the frame transformation node, which converts the $u, v$ coordinate frame to the camera frame and then to the world frame for the robot to pick up. Given the corresponding pose of targets, a collision-free trajectory for the robot is generated and executed by the C++ node. After picking  the bolt, the simulation adds a twisting motion to simulate unscrewing. Once all bolts in the first class are removed, the simulation proceeds to the second stage -- cable picking. With the bolts no longer locking down the cables, they are then ready to be picked. The whole workflow is demonstrated in Fig.~\ref{fig_Process}. Specifically, at the beginning of the simulation, A1 displays the first object that is targeted for disassembly (dark blue circle indicates the desired location to place the end effector of the robot), while A2 shows the arm moving to the location to pick it up. B1 shows the real-time disassembling process, where the perception node selects the next bolt while B2 shows the robot executing the task. In C1, it is displayed that there is only one bolt left, and  C2 shows a zoomed-in view of the gripper for picking. In D1, the transition between the first and second stages is shown as dark blue circle moving from the last bolt to the first cable whereas in D2 the gripper goes on to pick up the cables. After completing stage two, the modules are left to be removed. Since the modules require a vacuum gripper to pick them up, and the URDF does not support the dynamic changing of grippers, we switch the gripper by relaunching the corresponding URDF file and loading the vacuum gripper plugin. %This process assumes that h from where we left off in the previous stage, launching the environment with new vacuum grippers and remaining modules.
    In E1, the perception node selects the first module to pick, while E2 shows the robot successfully picking  this object. Finally, all the modules are removed. The objects MSD, positive bus bar, and negative bus are not implemented in this simulation, as the process for disassembling them  follows the same concepts as for the bolts and cables.
    The outcomes of the benchmark algorithm are presented in Table I. Notably, all  the assigned tasks were performed successfully. However, it is worth-noting that the analysis also reveals a prolonged  execution time for bolt disassembly, as generating a viable path took several attempts in some cases, highlighting a needed area for further enhancements over our C++ planner. While the current object detection algorithm is able to accurately detect all objects and their locations, the detection score is relatively low compared to that of larger objects. There is still room for improvement in detecting smaller objects with higher accuracy, which is a known challenge in computer vision. Nonetheless, the framework offers a comprehensive solution to automated battery recycling, and meanwhile there is certainly potential for further improvements and extensions, which will be discussed next.
 
\begin{table}[h]
\caption{Benchmark Algorithm Execution Results}
\vspace{-10pt}
\label{table_example}
\begin{center}
\begin{tabular}{c c c c}
\hline
 & Execution Time (s) & Detection Score (\%) & Success \\
\hline
Bolt 1 & 15.3 & 54 & Yes \\
Bolt 2 & 15.3 & 57 & Yes\\
Bolt 3 & 15.3 & 53 & Yes\\
Bolt 4 & 30.6 & 41 & Yes\\
Bolt 5 & 23.4 & 38 & Yes\\
Bolt 6 & 28.8& 35 & Yes\\
Cable 1 & 9.0& 53 & Yes\\
Cable 2 & 9.0& 47 & Yes\\
Module 1 & 9.0& 100 & Yes\\
Module 2 & 8.1& 100 & Yes\\
Module 3 & 18.0 & 100 & Yes\\
Module 4 & 9.0 & 100 & Yes\\
\hline
\end{tabular}
\end{center}
\end{table}
	
        \section{Extensions and Discussions} \label{sec_extension}
	The proposed framework is designed to be customizable and adaptable to accommodate different research needs. The framework is composed of various entities, each of which has been specifically designed to be highly interchangeable. The entities that can be exchanged include the battery model, robot model, perception algorithm, planning, and control.
    The battery model is currently developed for the module battery pack, but users have the option to create their own object of interest and integrate it into the Gazebo environment. This flexibility allows researchers to tailor the battery model to meet their specific research needs. A stable and computationally efficient way of creating simulations has been introduced in the framework. Working with stacked or complex objects no longer requires monotonous tuning of solid properties. Users can develop customized components based on their needs by following the similar steps discussed in this paper.
    
    The current implementation of the framework can use either the MoveIt OMPL library for planning or user-defined planner. We have provided an customized entry port for users who prefer to develop their own algorithm. This allows them to implement their own algorithm instead of relying on the Moveit planner, which may feel like a black box to some users.The framework's adaptability allows for the modification of the planning and control section. As long as the interface between the perception algorithm and the planning and control algorithm is maintained, the remaining components of the framework will continue to function as intended. This allows control researchers to concentrate on the development of the algorithm without having to delve into deep learning related areas.
    
    The current object detection model utilizes the SSD ResNet architecture, which has a good balance between speed and performance. However, users have the flexibility to select a model that meets their specific requirements and follow the training procedure in a similar manner. While the picture settings in the Gazebo world are ideal, the real world can present more variations, such as lighting, noise, and more. Moreover, to enhance the detection robustness, additional metrics can be incorporated. Data augmentation techniques, such as those described in \cite{chen2022deep}, can be applied to extend the dataset, enabling tests under a broader range of conditions. This will be crucial when detecting in the real world, where real hardware can be subject to noise. Overall, the modular design of the platform enables researchers to continue developing and improving their area of interest.
    
	\section{Conclusion} \label{sec_conclusion}
    This paper  presented a comprehensive simulation platform for research on robotic battery recycling. A generic EV battery pack consisting of 4 modules with inter-connected bolts and cables, a manipulator, and a Kinetic RGB-D camera were integrated as the main components of the simulated hardware system. A CAD model that depicts the EV battery pack was designed and imported into Gazebo for object interactions. Various procedures, including unscrewing, pulling, and lifting, were considered to imitate the key tasks involved in battery disassembly. Moreover, benchmark perception, planning, and control algorithms were provided to guide the manipulator in executing different disassembly tasks. The whole simulation platform was compact and efficient, and it can be used for further development, evaluation, and testing of robotic battery disassembly algorithms. Future works will include adding more simulated hardware components (e.g., different battery packs) to enhance the capabilities of the platform, exploring more efficient perception, planning and control algorithms, and developing multi-armed systems for collaborative disassembly.     

	\balance	
	\bibliographystyle{IEEEtran}
	\bibliography{IEEEabrv,reference}
\end{document}